\documentclass[sigconf]{acmart}

\usepackage{enumitem}
\usepackage{multirow}
\usepackage{colortbl}
\usepackage{algorithm}
\usepackage{algorithmic}

\AtBeginDocument{%
  }

\setcopyright{acmlicensed}
\copyrightyear{2018}
\acmYear{2018}
\acmDOI{XXXXXXX.XXXXXXX}
\acmConference[Conference acronym 'XX]{Make sure to enter the correct
  conference title from your rights confirmation email}{June 03--05,
  2018}{Woodstock, NY}
\acmISBN{978-1-4503-XXXX-X/2018/06}

\begin{document}

\title{Causal Negative Sampling via Diffusion Model for Out-of-Distribution Recommendation}

\author{Chu Zhao}
\affiliation{%
  \institution{Northeastern University}
  \city{Shenyang}
  \country{China}}
\email{chuzhao@stumail.neu.edu.cn}

\author{Enneng Yang}
\affiliation{%
  \institution{Northeastern University}
  \city{Shenyang}
  \country{China}}
  \email{ennengyang@stumail.neu.edu.cn}

\author{Yizhou Dang}
\affiliation{%
 \institution{Northeastern University}
 \city{Shenyang}
 \country{China}}
 \email{yzdang@stumail.neu.edu.cn
 }

\author{Jianzhe Zhao}
\affiliation{%
  \institution{Northeastern University}
    \city{Shenyang}
  \country{China}}
\email{zhaojz@swc.neu.edu.cn}

\author{Guibing Guo}
\authornote{Corresponding authors.}
\affiliation{%
 \institution{Northeastern University}
    \city{Shenyang}
  \country{China}}
\email{guogb@swc.neu.edu.cn}

\author{Xingwei Wang} 
\affiliation{%
 \institution{Northeastern University}
    \city{Shenyang}
  \country{China}}
\email{wangxw@mail.neu.edu.cn}

\renewcommand{\shortauthors}{Trovato et al.}

\begin{abstract}
Heuristic negative sampling enhances recommendation performance by selecting negative samples of varying hardness levels from predefined candidate pools to guide the model toward learning more accurate decision boundaries. However, our empirical and theoretical analyses reveal that unobserved environmental confounders (e.g., exposure or popularity biases) in candidate pools may cause heuristic sampling methods to introduce false hard negatives (FHNS). These misleading samples can encourage the model to learn spurious correlations induced by such confounders, ultimately compromising its generalization ability under distribution shifts.
To address this issue, we propose a novel method named \textbf{C}ausal \textbf{N}egative \textbf{S}ampling via \textbf{Diff}usion (CNSDiff). By synthesizing negative samples in the latent space via a conditional diffusion process, CNSDiff avoids the bias introduced by predefined candidate pools and thus reduces the likelihood of generating FHNS.
Moreover, it incorporates a causal regularization term to explicitly mitigate the influence of environmental confounders during the negative sampling process, leading to robust negatives that promote out-of-distribution (OOD) generalization.
Comprehensive experiments under four representative distribution shift scenarios demonstrate that CNSDiff achieves an average improvement of \textbf{13.96\%} across all evaluation metrics compared to state-of-the-art baselines, verifying its effectiveness and robustness in OOD recommendation tasks.
The code is available at \url{https://github.com/user683/CNSDiff}.

\end{abstract}




\maketitle

\section{Introduction}
Random negative sampling ~\cite{yang2024does,xu2022negative} selects negative examples randomly from the set of items that a user has not interacted with, which helps prevent model overfitting and refine the decision boundary. Therefore, it has been widely applied in recommendation systems~\cite{koren2009matrix,ma2024negative}. To further improve model performance, researchers have proposed various enhanced strategies over the past decade. For example, \textit{Predefined Negative Sampling}~\cite{song2015efficient,zhang2024practical} fixes negative examples during data preprocessing, while \textit{Popularity-based Negative Sampling}~\cite{yang2023hyperbolic,wang2019multi} selects negative samples based on item popularity, where more popular items are more likely to be chosen. Although these methods improve the quality of negative samples, they typically rely on a fixed hardness level, making it difficult to adapt to the model's varying discriminative ability at different training stages. Consequently, negative samples may be too challenging in the early stages, hindering convergence, or too simple in later stages, providing insufficient supervision signals, which limits the further improvement of model performance. 
To address this issue, recent studies have focused on \textit{Heuristic Negative Sampling}~\cite{huang2021mixgcf,lai2024adaptive,ma2023exploring}, which dynamically selects negative samples of varying hardness during training from predefined candidate pools. This approach continuously provides training signals that match the model's current capability, thereby accelerating convergence and improving recommendation performance.

\begin{figure}[t]
	\centering
	\begin{minipage}{0.48\linewidth}
		\centerline{\includegraphics[width=\textwidth]{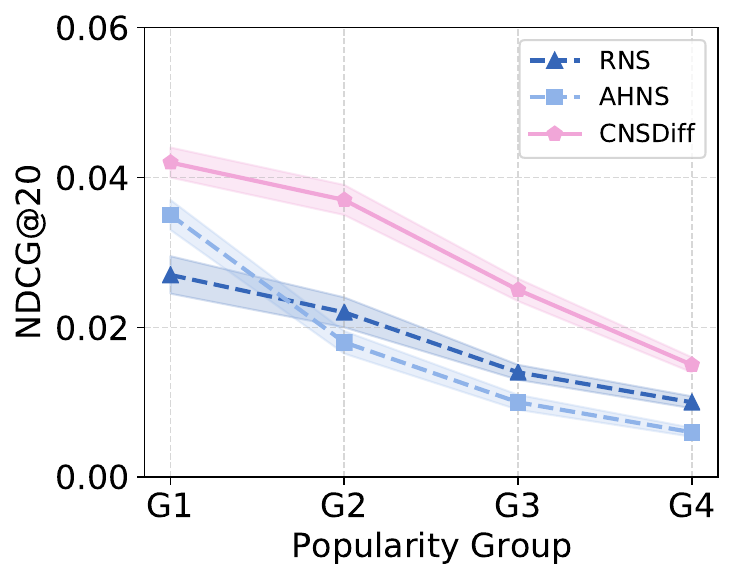}}
		\centerline{\footnotesize\;\;\;\;\;\;\;\; (a) }
	\end{minipage}
	\begin{minipage}{0.48\linewidth}
		\centerline{\includegraphics[width=\textwidth]{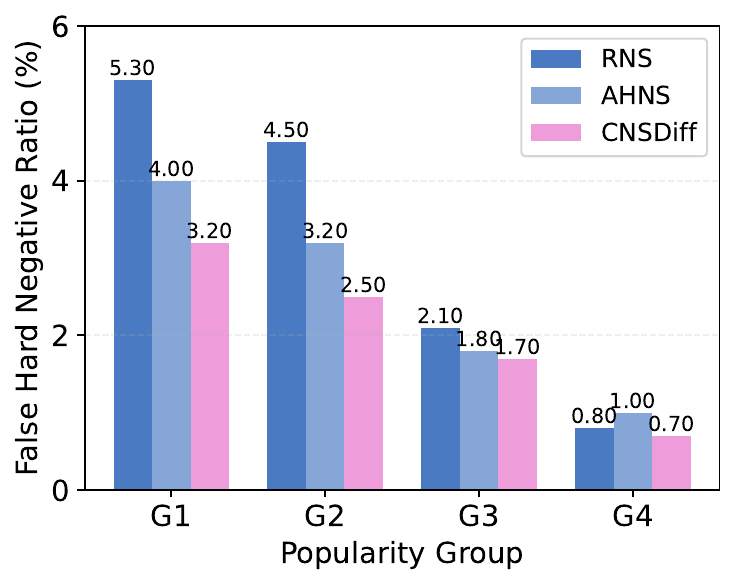}}
		\centerline{\footnotesize \;\;\;\;\;\;\;\;\;\;\;\;(b) }
	\end{minipage}
     \vspace{-0.3cm}
	\caption{Performance and false hard negative ratio Across different popularity groups.}
	\label{fig: empirical_study}
     \vspace{-0.5cm}
\end{figure}

Although heuristic negative sampling methods have shown strong performance in improving recommendation accuracy, our theoretical and empirical analyses reveal that their reliance on predefined candidate pools makes them vulnerable to unobserved confounders within the pool, which can lead to the inclusion of false hard negatives (FHNS) and ultimately undermine the model’s generalization ability. Such samples force the model to learn spurious correlations induced by environmental factors, ultimately degrading its generalization performance under distribution shifts.
Taking popularity as an example, low-exposure or niche items often fail to interact with users, but this does not imply that users are uninterested in them. Heuristic hard negative sampling, however, may mistakenly treat these potential positives as hard negatives that are highly similar to the user. For instance, a user may have a strong preference for an independent niche music brand, but due to its low popularity, it has never been recommended to them. Since the current embedding of such music brand is very close to this item, it is erroneously sampled as a hard negative. Compared to general false negatives, FHNS are more harmful: due to their high similarity to positive samples, the model is continuously “pushed away” from them during training, reinforcing incorrect decision boundaries. As a result, the model learns not the true preferences, but incorrectly associates “low-popularity, high-similarity items” with negative feedback. Such spurious correlations induced by popularity as a confounder severely undermine the causal robustness and out-of-distribution (OOD) generalization ability of the model.
In Section~3.1, we establish a structural causal model and further illustrate our experimental findings in Figure~\ref{fig: empirical_study}: the items are grouped into four buckets according to their popularity (from high to low), and we report the performance and the ratio of false negatives in each group. The results show that AHNS suffers from performance drops in certain popularity groups, even performing worse than random negative sampling; moreover, in high-popularity groups, heuristic negative sampling produces a significantly higher false hard negative ratio. This further verifies our hypothesis that environmental confounders induce FHNS, which in turn encourage the model to learn spurious correlations and harm its OOD generalization. In Figure~\ref{fig: empirical_study_temporal}, we further illustrate the performance variations across different temporal groups and the FHNS ratio, reaching the same conclusion.

To address the above issue, we propose a novel negative sampling strategy named \textbf{CNSDiff}. Unlike traditional heuristic methods that rely on predefined candidate pools, CNSDiff takes the positive item embeddings as conditional input and leverages a  diffusion model to progressively perturb them in the latent space, generating negative samples with varying hardness levels.
The core intuition behind this strategy is that the initial diffusion steps preserve the semantic structure of the positive sample, while the subsequent noise injection and denoising process gradually move it away from the original distribution. This results in \textbf{high-quality hard negatives} that are close to but distinct from the positive items.
We further provide the formal mathematical definition in Section 3.1.
Furthermore, as illustrated in Figure \ref{fig: causal} (a), we construct a structural causal model (SCM) to characterize the data generation mechanism and analyze how environmental confounders may introduce spurious hard negatives, thus impairing the model's OOD generalization capability. Based on this, we design a causal regularization term that leverages the backdoor criterion to identify and block the causal path between environmental confounders and the target variable, thereby enforcing distributional consistency of the diffusion model across different environments. This design enables CNSDiff to effectively mitigate FHNS induced by confounding factors, resulting in more robust negative sample generation. The main contributions of this work are shown as follows:
\begin{itemize}[leftmargin=0pt]
    \item We construct a structural causal model to reveal that environmental confounders can introduce FHNS during heuristic negative sampling, leading the model to learn spurious environment-dependent correlations and degrading its OOD generalization performance.
    \item We propose a diffusion-based negative sampling method with causal regularization, which utilizes the backdoor criterion in the negative sample generation process to identify and block the causal paths from environmental confounders to the target variable. This approach enforces distributional consistency of the diffusion model across environments, significantly enhancing the robustness of negative samples and the model's generalization capability.
    \item Extensive experiments across various distribution shift scenarios demonstrate that our proposed method consistently outperforms baselines in terms of OOD generalization performance. 
\end{itemize}

\section{Preliminary} 
\subsection{Problem Formulation}
In recommender systems, let the user set be $ \mathcal{U} = \{u_i\}_{i=1}^{M} $, the item set be $ \mathcal{V} = \{v_i\}_{i=1}^{N} $, and the user-item interaction matrix be $ R \in \mathbb{R}^{M \times 
N} $, where $ R_{uv} = 1 $ indicates an interaction between user $ u $ and item $ v $, and $ R_{uv} = 0 $ otherwise. Models typically learn an encoder function $f_{\theta}(\cdot)$ to map users and items into low-dimensional embeddings and predict user preferences based on their similarity. A widely used optimization objective is the Bayesian Personalized Ranking (BPR) loss:
\begin{equation}
    \mathcal{L}_{\text{BPR}} = \sum_{(u,v) \in R} -\log \sigma(s(u,v) - s(u,v^-)),
    \label{eq: bpr}
\end{equation}
where $ s(u,v) $ denotes the predicted score of user $ u $ for item $ v $, $ \sigma $ is the sigmoid function, and $ v^- $ represents a negatively sampled item not interacted with by the user. BPR trains the model by contrasting positive and negative samples to capture user preference. Existing methods often adopt heuristic negative sampling strategies, selecting negatives from a candidate item pool either randomly or by rule. However, such approaches are prone to introducing false hard negatives (FHNS). Through empirical analysis, we demonstrate that FHNS can amplify spurious correlations caused by environmental confounders, ultimately harming the model’s generalization to OOD scenarios. Therefore, this work aims to design a novel negative sampling strategy to mitigate the impact of FHNS and enhance the model’s OOD generalization performance.

\subsection{Diffusion Model}
Diffusion model~\cite{croitoru2023diffusion, zhao2024denoising} is a class of deep generative models that have achieved significant success in computer vision due to their high-quality generation, training stability, and solid theoretical grounding. They consist of two key phases: a \textbf{forward process}, which progressively adds Gaussian noise to data $ x_0 \sim p(x_0) $, here, $p(x_0)$ denotes the \textbf{true data distribution}, and a \textbf{reverse process}, which learns to recover the original data.
In the forward process, noise is added step-by-step:
\begin{equation}
q(x_t | x_{t-1}) = \mathcal{N}(x_t; \sqrt{1 - \beta_t} x_{t-1}, \beta_t \cdot \mathbf{I}),
\end{equation}
where $ \beta_t \in (0,1) $ controls the noise level at step $ t $. $\textbf{I}$ is the identity matrix. The notation $\mathcal{N}(\cdot; \mu, \Sigma)$ denotes a multivariate Gaussian distribution with mean $\mu$ and covariance $\Sigma$. Thus, $x_t$ is sampled by scaling $x_{t-1}$ and adding zero-mean Gaussian noise with variance $\beta_t$, making the sample gradually more noisy. Using the reparameterization trick, we have:
\begin{equation}
q(x_t | x_0) = \mathcal{N}(x_t; \sqrt{\bar{\alpha}_t} x_0, (1 - \bar{\alpha}_t)\cdot\mathbf{I}),
\end{equation}
where $ \bar{\alpha}_t = \prod_{i=1}^t (1 - \beta_i) $.
The reverse process learns a denoising model $ p_\theta(x_{t-1} | x_t) $ parameterized by neural networks:
\begin{equation}
p_\theta(x_{t-1} | x_t) = \mathcal{N}(x_{t-1}; \mu_\theta(x_t, t), \Sigma_\theta(x_t, t)).
\end{equation}
Training minimizes the variational lower bound (VLB):
\begin{equation}
\mathcal{L}_{\text{VLB}} = \mathbb{E}_{q} \left[ \sum_{t=1}^T \mathrm{KL}(q(x_{t-1} | x_t, x_0) \| p_\theta(x_{t-1} | x_t)) \right] - \log p_\theta(x_0 | x_1),
\end{equation}
which is often simplified to a mean squared error (MSE) loss on noise prediction:
\begin{equation}
\mathcal{L}_{\text{sampling}} = \mathbb{E}_{t,x_0,\epsilon} \left[ \| \epsilon - \epsilon_\theta(\sqrt{\bar{\alpha}_t} x_0 + \sqrt{1 - \bar{\alpha}_t} \epsilon, t) \|^2 \right].
\end{equation}
This reformulation not only improves training stability but also enables efficient optimization using standard regression techniques. In our setting, the diffusion model enables controllable synthesis of negative samples by varying the diffusion step $T=\{0\le t\le T\}$. Smaller $T$ yields harder negatives closer to the original item $x_0$, while larger $T$ produces easier negatives with more noise. This provides a principled way to generate negatives of varying hardness in continuous space, without relying on heuristic candidate sets.

\begin{figure}[t]
	\centering
	\begin{minipage}{1\linewidth}
\centerline{\includegraphics[width=\textwidth]{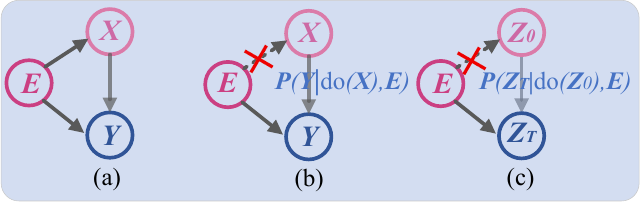}}
	\end{minipage}
    \vspace{-0.2cm}
	\caption{Causal Graph Analysis Model. E represents unobserved environmental factors, X denotes the input (i.e., user-item pairs), and Y is the model’s predicted output.}
     \label{fig: causal}
      \vspace{-0.5cm}
\end{figure}

\section{Methodology} 
We now present our analysis and the proposed model. In Section~3.1, we construct a causal structural model (SCM) to characterize the data generation process, illustrating how false hard negatives (FHNS) are introduced and how they amplify spurious correlations that undermine out-of-distribution generalization. Section~3.3 details the diffusion-based mechanism for synthesizing negative samples with controllable hardness. In Section~3.4, we introduce a causal regularization term that mitigates the influence of confounding factors during negative sampling. Finally, we provide a time complexity analysis and discuss why CNSDiff is effective.

\subsection{Causal Analysis for Negative Sampling}
To investigate the impact of FHNS on the out-of-distribution (OOD) generalization of models, we conduct a causal analysis to uncover the dependencies among key variables in the recommendation process. Our analysis reveals that FHNS amplify spurious correlations introduced by environment-specific factors, thereby degrading model robustness under distribution shifts.
We construct the SCM (Figure~\ref{fig: causal}), where $E$ denotes the environment, $X$ represents the user-item input pair, and $Y$ is the model’s prediction. The causal dependencies are characterized as follows:

\begin{itemize}[leftmargin=0pt]
    \item $\mathbf{E} \rightarrow \mathbf{X}$: The environment influences the exposure distribution $p(X \mid E)$, introducing biases such as popularity or temporal shift.
    \item $\mathbf{X} \rightarrow \mathbf{Y}$: The model deterministically maps $X$ to prediction via $Y = f_\theta(X)$.
    \item $\mathbf{E} \rightarrow \mathbf{Y}$: Although not directly modeled, this path is implicitly introduced by environment-biased negative sampling during training.
\end{itemize}

We first provide the mathematical definition of FHNS:
\begin{definition}[\textbf{False Hard Negative Sample}]
In the presence of environmental confounders, certain user-preferred items may have low exposure probability and remain unobserved. Let $Y_{u,v} \in \{0,1\}$ denote the true preference of user $u$ for item $v$, and $\phi(v, e)$ denote the exposure probability of item $v$ under environment $e$. A sample $(u,v)$ is defined as a \textbf{False Hard Negative Sample (FHNS)} if:
\begin{equation}
\text{FHNS}_{u,v} := \left\{ \tilde{Y}_{u,v} = 0 \;\middle|\; Y_{u,v} = 1,\; \phi(v, e) \ll 1,\; \text{sim}(z_u, z_v) \ge \tau \right\},
\end{equation}
where $\tilde{Y}_{u,v} = 0$ is the sampled label observed as a negative, $Y_{u,v} = 1$ indicates a true preference of user $u$ for item $v$, $\phi(v, e) \ll 1$ denotes a low exposure probability of item $v$ under environment $e$, and $\text{sim}(z_u, z_v) \ge \tau$ implies high semantic similarity between the user and item embeddings, exceeding a threshold $\tau$. Such false negatives mislead the model by incorrectly penalizing relevant items, potentially amplifying spurious correlations and impairing generalization under distribution shifts.
\end{definition}
These FHNS are often indistinguishable from true positives in the representation space, thereby introducing high-magnitude noise during training. The BPR loss in Eq.~(\ref{eq: bpr}) can be further expressed as:
\begin{equation}
\mathcal{L}_{\text{BPR}}(\theta) = -\mathbb{E}_{(u, v, v^-) \sim \mathcal{D}} \left[\log \sigma\left(f_\theta(u, v) - f_\theta(u, v^-)\right) \right],
\end{equation}
when the sampled negative $v^-$ is actually a false hard negative (i.e., $\tilde{Y}_{u,v^-} = 0$ but $Y_{u,v^-} = 1$) and $\mathcal{D}$ is the dataset, the model is misled to push a truly relevant item away from the user in the embedding space. This misoptimization distorts the learned decision boundary and significantly degrades recommendation performance, especially under distribution shift.
We define the \textbf{environment-conditioned false hard negative rate} as:
\begin{equation}
\eta(e) := P(\tilde{Y} = 0 \mid Y = 1, E = e),
\end{equation}
which captures the probability that true positives are mislabeled as negatives under environment $e$. As a result, the model learns a weighted version of the true label distribution:
\begin{equation}
f_\theta(X) \approx \mathbb{E}[\tilde{Y} \mid X = x ] = \mathbb{E}_E \left[ P(Y = 1 \mid X, E) \cdot (1 - \eta(E)) \right],
\end{equation}
This implies a spurious dependence of prediction on $E$ even when $X$ is fixed, leading to overfitting to environment-specific artifacts and degraded OOD generalization.  We further elaborate in Theorem~\ref{TH: FHNS} how FHNS amplify spurious correlations.
\begin{theorem}[\textbf{False Hard Negatives Amplify Environment-Induced Spurious Correlations}]
\label{TH: FHNS}
Suppose the false hard negative rate $ \eta(e) $ varies across environments $ e \in E $, and the model $ f_\theta(x) $ is differentiable with respect to a latent variable $ z_E $ aligned with the environmental variable $ E $. If the test environment distribution $ P_{\text{test}}(E) $ differs from the training distribution $ P_{\text{train}}(E) $, then the generalization gap admits the following lower bound:
\begin{equation}
  \mathcal{L}_{\text{test}}(f_\theta) - \mathcal{L}_{\text{train}}(f_\theta) \geq \gamma \cdot \mathrm{TV}(P_{\text{train}}(E), P_{\text{test}}(E)) + \varepsilon_{\text{inv}}.  
\end{equation}
\end{theorem}
\noindent where $ \mathrm{TV}(P_{\text{train}}(E), P_{\text{test}}(E)) = \frac{1}{2} \int |P_{\text{train}}(e) - P_{\text{test}}(e)|\,de $ is the total variation distance between the training and test environment distributions,  
$ \gamma = \mathbb{E}_X \left[ \left\| \frac{\partial f_\theta(X)}{\partial z_E} \right\| \right] \cdot \mathrm{Var}_e(\eta(e)) $ quantifies the model's sensitivity to environment-specific false negative perturbations,  
and $ \varepsilon_{\text{inv}} $ captures the irreducible invariant risk.  
A complete proof is provided in Appendix~\ref{pf: T_3.1}.

\subsection{Negative Sampling via Diffusion}
Inspired by \cite{nguyen2024diffusion}, we take the generated item embeddings as the input to the diffusion model and extract the embeddings from multiple time steps to implement a hierarchical generative negative sampling strategy. Specifically, CNSDiff controls the difficulty of negative samples by adjusting the diffusion time step $ T $. For instance, when $ T = 0 $, the generated embeddings are from the final stage of the diffusion process and are the hardest to distinguish from positive samples in the latent space, thus regarded as the \emph{hardest negatives}. In contrast, embeddings generated at earlier diffusion steps (i.e., larger $ T $) tend to be easier to distinguish from positives. In general, the time step $ T $ serves as a proxy for the hardness of negative samples: the smaller the $ T $, the more challenging the negative samples.
Specifically, taking graph-based collaborative filtering as an example, given a user-item interaction pair $\langle u, i \rangle$, we first construct an interaction graph $ G $. Then, a graph convolutional network is employed to obtain the user embeddings $z_u$ and item embeddings $z_v$, respectively. The positive item embeddings $z_v$ are subsequently used as the input to the diffusion model. The overall processing pipeline is illustrated as follows:
\begin{equation}
z_v \rightarrow \mathbf{z}_0 \xrightarrow{\phi}\cdots \mathbf{z}_t \xrightarrow{\psi} \hat{\mathbf{z}}_{t-1} \cdots \xrightarrow{\psi} \hat{\textbf{z}}_0,
\label{eq: diffusion_process}
\end{equation}
where $\phi$ represents injecting noise into the input embedding, while $\psi$ removes the noise through optimization. $\hat{\textbf{z}}_0$ represents the denoised embedding.
CNSDiff controls both the quantity and hardness of the generated negative samples by selecting outputs from specific diffusion time steps between $0$ and $T$. The embedding at time step $T$ can be obtained as follows:
\begin{equation}
    p_{\theta}(x_{t-1}|x_t) = \frac{1}{\sqrt{\alpha_t}} \left ( x_t - \frac{1 - \alpha_t}{\sqrt{1 - \bar{\alpha}_t}} \right) + \sqrt{\beta_t} \cdot \mathbf{I}.
\end{equation}
Through the above sampling strategy, CNSDiff constructs a set of generated negative samples with different levels of \emph{hardness}, denoted as $\{ h_t \mid 0 < t < T \}$. In practice, the choice of diffusion time steps has a significant impact on both the quality of negative samples and training efficiency. Smaller time steps (i.e., those closer to $t=0$) tend to produce more challenging negatives, but at the cost of increased computational overhead. In contrast, larger time steps may generate easily distinguishable negatives that provide limited learning signals for the model.
To strike a balance between training efficiency and the diversity of generated negative samples, we adopt a uniformly spaced sampling strategy across the diffusion process. Instead of using all intermediate steps, we select a subset of representative time steps to generate negatives, thereby covering samples with different levels of semantic hardness while reducing computational cost. We define the set of sampling steps as:
\begin{equation}
\mathcal{D}_{h_t} = \{h_t \mid t = t_0 + k \cdot s,\ t \leq T,\ k \in \mathbb{N}\},
\end{equation}
where $t_0$ denotes the initial step, $s$ is the stride (i.e., interval between sampled steps), and $T$ is the total number of diffusion steps.  This strategy ensures both sampling efficiency and sufficient coverage of varying negative sample difficulty.
\begin{figure}[t]
    \centering
    \includegraphics[width=0.48\textwidth]{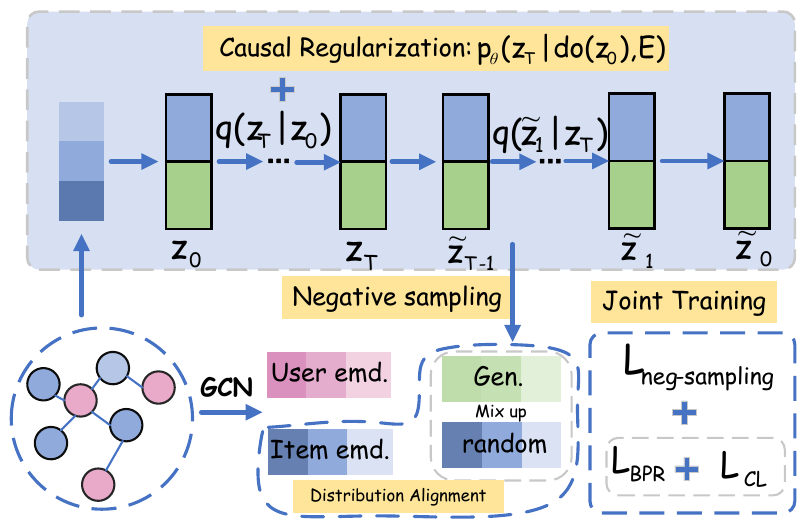}
    \caption{
        Illustration of the proposed CNSDiff architecture. The model takes the item embeddings as input and synthesizes a set of robust negative samples with varying levels of hardness via diffusion-based causal regularization. These samples are then integrated with random negatives through mixup and aligned semantically, enabling a joint training framework that enhances the generalization.
    }
     \vspace{-0.5cm}
    \label{fig: CNSDiff}
\end{figure}
\subsection{Optimization with Causal Regularization}
As discussed in Section 3.1, FHNS can amplify spurious correlations introduced by environmental variables, thereby undermining the model's out-of-distribution generalization ability. The aforementioned diffusion-based negative sampling methods do not explicitly address environmental confounders, and thus struggle to achieve robust generalization.
Inspired by recent advances, a feasible solution is to perform deconfounded modeling by estimating the interventional distribution:
$p_\theta(Y \mid \text{do}(X), E)$,
which aims to block the dependency path from the environment variable $E$ to the input feature $X$, and consequently uncover stable and transferable causal relationships between $X$ and $Y$.
In the following, we explore how to explicitly learn causally invariant user preference representations during the negative sample generation process using diffusion models, with the goal of enhancing generalization under distribution shifts. Given an initial embedding $z_0$, the generative process under an unobserved environment variable $E = e$ can be modeled as a conditional diffusion chain:
\begin{equation}
  p_\theta(z_T \mid z_0, E = e) = \prod_{t=0}^{T-1} p_\theta(z_{t+1} \mid z_t, e),  
\end{equation}
where $z_0$ denotes the starting point of the diffusion process and $T$ is the total number of diffusion steps. The corresponding log-likelihood can be expressed as:
\begin{equation}
 \log p_\theta(z_T \mid z_0, E = e) = \sum_{t=0}^{T-1} \log p_\theta(z_{t+1} \mid z_t, e).   
\end{equation}
To eliminate the influence of unobserved confounding factors on negative sample generation and simultaneously capture user preferences that are invariant across environments, an ideal approach is to adopt deconfounder learning. This can be formulated as modeling the interventional distribution: $
p_\theta(z_T \mid \operatorname{do}(z_0), E=e)$,
where the do-operator removes spurious associations induced by environment-dependent confounding. We present the following theory directly.
\begin{theorem}For a given diffusion model $p_\theta(z_T \mid z_0, e)$, which models the generation of latent representations conditioned on both the initial embedding and environment variable, the deconfounded learning objective admits the following variational lower bound:
\begin{multline}
\log p_\theta(z_T \mid \operatorname{do}(z_0), E=e) 
\geq \\ \sum_{t=0}^{T-1} \mathbb{E}_{q_\phi(e \mid z_t)} 
\left[ \log \frac{p_\theta(z_{t+1} \mid z_{t}, e) \, p_0(e)}
{q_\phi(e \mid z_{t})} \right],
\end{multline}
where $q_\phi(e \mid z_{t})$ is a variational approximation of the posterior over environment variables, and $p_0(e)$ denotes the prior over environments.
\end{theorem}
\noindent The proof is deferred to Appendix~\ref{proof：Variational}. Notably, the variational objective can be interpreted as an importance reweighting mechanism, where the ratio. 
$\frac{p_0(e)}{p(e \mid z_t)}$ 
adjusts for the imbalance in the distribution of the environment variable. This reweighting downplays dominant environments and emphasizes under-represented ones, thereby mitigating the observation bias induced by skewed or limited training data. Consequently, the model learns more stable and invariant transitions. 
$ z_t \rightarrow z_{t+1}$ generalizes beyond the training distribution. Overall, this variational formulation facilitates effective deconfounding in generative representation learning by explicitly modeling and correcting for latent environmental confounders.
Based on Eq.~(17), we propose a diffusion-based negative sampling objective with causal regularization:
\begin{equation}
\mathcal{L}_{\text{neg\_sampling}} = \mathcal{L}_{\text{sampling}} + \lambda_1 \cdot \log p_\theta(z_T|\text{do}(z_0),E),
\end{equation}
where $ \mathcal{L}_{\text{sample}} $ denotes the sampling loss from generated negatives, $ \lambda_1 $ is a penalty coefficient. This regularization encourages the model to eliminate spurious correlations introduced by confounding factors and to learn stable representations across varying environments. We provide the full derivation of the causal regularization term in Appendix~\ref{proof: Causal}.

\subsection{Joint Training}
After obtaining the set of negative samples $\mathcal{D}_h$ generated by the diffusion model, we follow previous work and select the most informative negative sample by computing the similarity between each generated item and the user embedding. Specifically, we choose the generated sample with the highest inner product score with the user embedding $z_u$:
\begin{equation}
    e_h = \arg\max_{h_t \in \mathcal{D}_h} (z_u \cdot h_t),
\end{equation}
where $h_t$ denotes a generated candidate from $\mathcal{D}_h$. This ensures that $e_h$ is the most attractive (i.e., hard) negative sample for the user, which enhances the difficulty and informativeness of the learning signal.
To improve training stability and generalization, we further introduce a progressive interpolation strategy that mixes the selected hard negative $e_h$ with a randomly sampled negative $e_r$ to construct the final negative embedding:
\begin{equation}
    \tilde{e} = \alpha \cdot e_r + \beta \cdot e_h,
\end{equation}
where $\alpha$ and $\beta$ are interpolation weights, and dynamically evolve with training process. In the early stages of training, we assign a higher weight to randomly sampled negatives ($\alpha > \beta$) to encourage diversity and exploration. As training progresses, we gradually increase the contribution of the generated negatives ($\beta > \alpha$), allowing the model to focus on more challenging and informative examples.
This curriculum-inspired mixing strategy balances easy and hard samples over time, improving both robustness and OOD generalization.
Furthermore, to enhance the discriminability of the generated negatives in the item embedding space, we adopt a contrastive learning objective. The contrastive loss is formally defined as follows:
\begin{equation}
    \mathcal{L}_{CL} = -\log \frac{\exp\left( \frac{\mathbf{z}_0 \cdot \mathbf{z}_j}{\tau} \right)}{\exp\left( \frac{\mathbf{z}_0 \cdot \mathbf{z}_j}{\tau} \right) + \sum_{k \neq j} \exp\left( \frac{\mathbf{z}_0 \cdot \mathbf{\tilde e}_k}{\tau} \right)},
\end{equation}
where $\mathbf{z}_j$ denotes the embeddings of the anchor and its positive counterpart, respectively, and $\tau$ is a temperature parameter that controls the concentration of similarity scores. The negative samples in the denominator include both generated and randomly sampled items. We adopt a joint optimization framework to train the model:
\begin{equation}
    \mathcal{L}_{\text{total}} = \mathcal{L}_{\text{BPR}} + \lambda_2 \cdot \mathcal{L}_{\text{neg\_sampling}} + \lambda_3 \cdot \mathcal{L}_{\text{CL}},
\end{equation}
where $\lambda_2$ and $\lambda_3$ are hyperparameters that control the contribution of the negative sampling and contrastive loss terms, respectively. The model architecture is illustrated in Figure~\ref{fig: CNSDiff}, and the algorithm pseudocode is presented in Algorithm~\ref{alg:CNSDiff}.

\subsection{Time Complexity}
In LightGCN, the time complexity of neighborhood aggregation and embedding update is $\mathcal{O}(K\cdot |E|\cdot d)$. $K$ is the number of graph convolution layers, $E$ is the nunber of edges, and $d$ is the dimensionality of the embeddings.  
With the introduction of the diffusion-based for negative sampling, each positive sample requires the generation of $M$ negative samples via a diffusion process with $T$ steps, where each step incurs a computation cost of $\mathcal{O}(d^2)$.  
Therefore, the overall time complexity of negative sampling becomes $\mathcal{O}(N_{\text{pos}} \cdot M \cdot T \cdot d^2)$ where $N_{\text{pos}}$ is the number of positive user-item pairs in the training data. 
The total training complexity is:
$\mathcal{O}(K\cdot |E|\cdot d + N_{\text{pos}} \cdot M \cdot T \cdot d^2)$.
Additionally, we benchmark the per-epoch runtime of CNSDiff against other state-of-the-art negative sampling methods, as shown in Table~\ref{tab: time}.  
CNSDiff demonstrates superior efficiency compared to MixGCF and AHNS, highlighting its scalability during training.

\begin{table*}[t]
\renewcommand{\arraystretch}{1.1}
\centering
\caption{Performance comparison of CNSDiff with state-of-the-art baselines under distribution shift scenarios on the Yelp2018, KuaiRec, and Food datasets. We report Recall@K (R@K) and NDCG@K (N@K) as evaluation metrics. The best results are highlighted in bold, and the second-best results are underlined.}
\setlength{\tabcolsep}{0.8mm}
\begin{tabular}{c|cccc|cccc|cccc}
\hline
 \multirow{2}{*}{\textbf{Methods}}& \multicolumn{4}{c|}{\textbf{Yelp2018}} & \multicolumn{4}{c|}{\textbf{KuaiRec}} & \multicolumn{4}{c}{\textbf{Food}} \\
\cline{2-13}
 & R@10 & N@10 & R@20 & N@20 & R@10 & N@10 & R@20 & N@20 & R@10 & N@10 & R@20 & N@20 \\
\hline
RNS ( \textcolor{blue}{SIGIR'20} )         & 0.0033 & 0.0020 & 0.0079 & 0.0036 & 0.0742 & 0.5096 & 0.1120 & 0.4268 & 0.0233 & 0.0182 & 0.0404 & 0.0242 \\
DNS(M, N) ( \textcolor{blue}{WWW'23} )      & 0.0014 & 0.0006 & 0.0128 & 0.0036 & 0.0529 & 0.3111 & 0.0829 & 0.2658 & 0.0146 & 0.0107 & 0.0274 & 0.0150 \\
MixGCF ( \textcolor{blue}{KDD'22} )        & 0.0027 & 0.0016 & 0.0078 & 0.0034 & 0.0541 & 0.3341 & 0.0963 & 0.2856 & 0.0162 & 0.0113 & 0.0295 & 0.0158 \\
AHNS ( \textcolor{blue}{AAAI'24} )         & 0.0008 & 0.0005 & 0.0032 & 0.0013 & 0.0526 & 0.3097 & 0.0829 & 0.2651 & 0.0151 & 0.0101 & 0.0304 & 0.0154 \\
DMNS ( \textcolor{blue}{WWW'24} )   & 0.0006 & 0.0003 & 0.0012 & 0.0005 & 0.0760 & 0.5078 & 0.1074  & 0.4300 & 0.0275 & 0.0201 & 0.0463 & 0.0272  \\
InvCF ( \textcolor{blue}{WWW'23} )         & 0.0016 & 0.0008 & 0.0013 & 0.0008 & 0.0871 & 0.2021 & 0.1024 & 0.2083 & 0.0030 & 0.0012 & 0.0033 & 0.0013 \\
AdvInfoNCE ( \textcolor{blue}{NIPS'23} )   & 0.0047 & 0.0024 & 0.0083 & 0.0038 & 0.0744 & 0.4109 & 0.0985 & 0.4020 & 0.0227 & 0.0135 & 0.0268 & 0.0159 \\
CDR ( \textcolor{blue}{TOIS'23} )          & 0.0011 & 0.0006 & 0.0016 & 0.0008 & 0.0570 & 0.2630 & 0.0860 & 0.2240 & 0.0260 & 0.0195 & 0.0412 & 0.0254 \\
DR-GNN ( \textcolor{blue}{WWW'24} )        & 0.0044 & 0.0029 & 0.0076 & 0.0041 & 0.0808 & 0.5326 & 0.1066 & 0.4511 & 0.0246 & 0.0205 & 0.0436 & 0.0272 \\
DiffRec ( \textcolor{blue}{SIGIR'23} )     & 0.0007 & 0.0004 & 0.0016 & 0.0007 & 0.0011 & 0.0158 & 0.0023 & 0.0128 & 0.0036 & 0.0029 & 0.0078 & 0.0050 \\
HDRM ( \textcolor{blue}{WWW'25} )  & 0.0025 & 0.0017 & 0.0046 & 0.0024 & 0.0410 & 0.2742 & 0.0645 & 0.2332 & 0.0141 & 0.0117 & 0.0257 & 0.0165         \\
CausalDiff ( \textcolor{blue}{WWW'25} )   & \underline{0.0077} & \underline{0.0049} & \underline{0.0130} & \underline{0.0068} &  \underline{0.0880}    & \underline{0.5410}     & \underline{0.1132} & \underline{0.4523}  & \underline{0.0285} & \underline{0.0209} & \underline{0.0471} & \underline{0.0278}   \\
\rowcolor{gray!30}\textbf{CNSDiff (Ours)} & \textbf{0.0101} & \textbf{0.0069} & \textbf{0.0163} & \textbf{0.0091} & \textbf{0.0909} & \textbf{0.5572} & \textbf{0.1384} & \textbf{0.4678} & \textbf{0.0304} & \textbf{0.0225} & \textbf{0.0496} & \textbf{0.0290} \\
\hline
\textbf{Improv. \%}& +31.69\% & +40.82\% & +25.38\% & +33.82\% & +3.59\% & +2.99\% & +2.23\% & +3.42\%  & +6.31\% & +7.66\% & +5.31\% & +4.32\% \\
\hline
\end{tabular}
\label{tab: main-results}
\end{table*}

\subsection{Discussion}
This section analyzes \textbf{why the negative samples synthesized by CNSDiff can improve the model's OOD generalization ability}.
Existing studies~\cite{yang2020understanding,nguyen2024diffusion} have pointed out that an ideal negative sample distribution $p_{v^-}$ should exhibit a \textit{sub-linear positive correlation} with the positive sample distribution $p_{v}$, i.e., it should satisfy $ p_{v^-} \propto p_v^\alpha $, where $ 0 < \alpha < 1 $. Furthermore, existing work~\cite{nguyen2024diffusion} theoretically proves that under a properly designed diffusion modeling setup, the distribution of negative samples naturally adheres to this sub-linear positivity principle. As a result, negative samples synthesized by diffusion models tend to exhibit strong training performance and generalization ability.
However, it is important to note that this theoretical foundation relies on the assumption of a stable data distribution, which may not hold in real-world scenarios with distribution shift. As demonstrated by the results of DMNS in Table~\ref{tab: main-results} and the ablation setting CNSDiff w/o All in Table~\ref{tab: ab_test}, failure to account for latent \textit{environmental confounders} can lead to diffusion-generated negative samples being influenced by spurious factors, thereby impairing OOD performance.
To address this issue, we extend the above theoretical framework by introducing a \textit{causal regularization} term, which explicitly enforces environment-invariance during the diffusion process. This constraint effectively mitigates the impact of confounders and further enhances the robustness and generalization capability of the model under distribution shifts.

 \section{Experiment}
We conduct extensive experiments to evaluate the effectiveness of CNSDiff and aim to address the following key scientific questions:
\textbf{(RQ1)} How does CNSDiff perform compared to state-of-the-art (SOTA) negative sampling methods? Can it enhance model generalization in the face of distributional shifts?
\textbf{(RQ2)} How do the individual components of CNSDiff contribute to the performance?
\textbf{(RQ3)} How do different hyperparameter settings of CNSDiff affect its OOD generalization capability?
\textbf{(RQ4)} What is the potential of CNSDiff in mitigating the challenges of false hard negative samples (FHNS) problem?

\textbf{Datasets}. 
Following prior work [12,13], we evaluate the performance of CNSDiff under three representative distribution shift scenarios: popularity shift, temporal shift, and exposure shift. Experiments are conducted on three widely used benchmark datasets: Food, KuaiRec, and Yelp2018.  Comprehensive details about the datasets and preprocessing procedures are provided in Appendix~\ref{ap: dataset}.

\textbf{Baselines}. 
We compare CNSDiff with three categories of SOTA baselines:
(1) \textbf{Negative sampling methods}: RNS \cite{koren2009matrix}, DNS (M, N) \cite{shi2023theories}, MixGCF~\cite{huang2021mixgcf}, AHNS~\cite{lai2024adaptive}, and DMNS~\cite{nguyen2024diffusion}.
(2) \textbf{Distribution shift-aware methods}: InvCF~\cite{zhang2023invariant}, AdvInfoNCE~\cite{zhang2024empowering}, CDR ~\cite{wang2023causal}, and DR-GNN~\cite{wang2024distributionally}.
(3) \textbf{Diffusion-based methods}: DiffRec~\cite{wang2023diffusion}, HDRM~\cite{yuan2025hyperbolic}, and CausalDiff~\cite{zhao2025graph}.
Detailed descriptions of these baselines are provided in Appendix~\ref{ap: dataset}.  
The hyperparameter configurations for CNSDiff are outlined in Appendix~\ref{ap: Hyper}.

\begin{figure*}[t]
	\centering
	\begin{minipage}{0.24\linewidth}
	\centerline{\includegraphics[width=\textwidth]{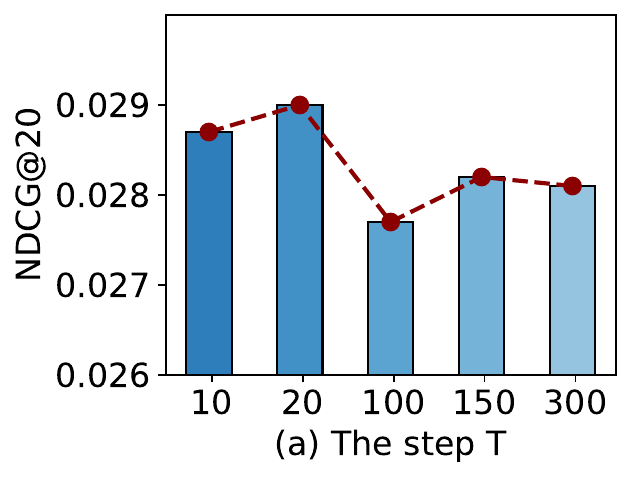}}
	\end{minipage}
	\begin{minipage}{0.24\linewidth}
	\centerline{\includegraphics[width=\textwidth]{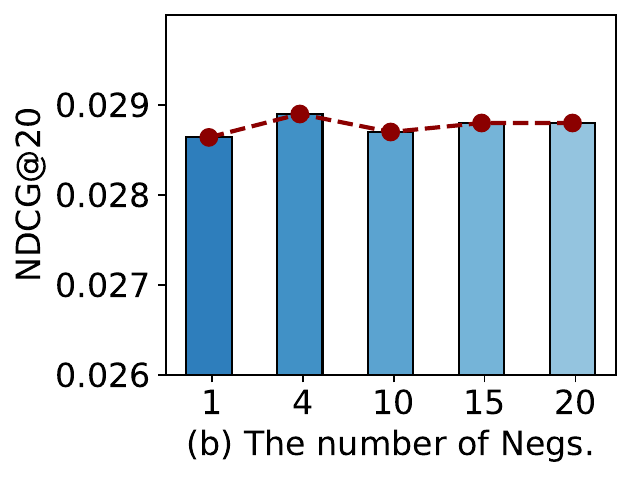}}
	\end{minipage}
	\begin{minipage}{0.24\linewidth}
	\centerline{\includegraphics[width=\textwidth]{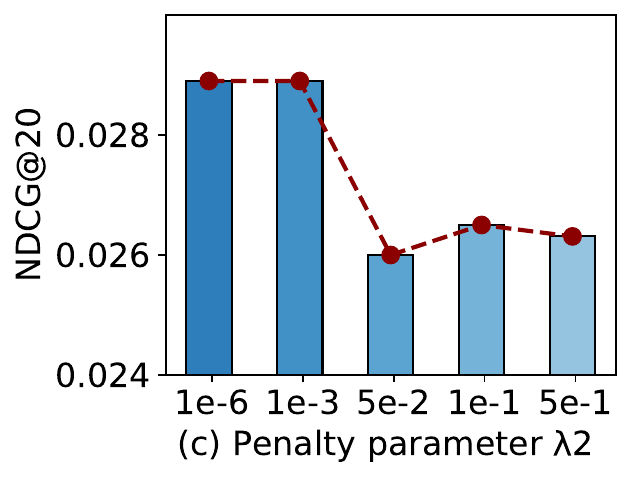}}
	\end{minipage}
    \begin{minipage}{0.24\linewidth}
	\centerline{\includegraphics[width=\textwidth]{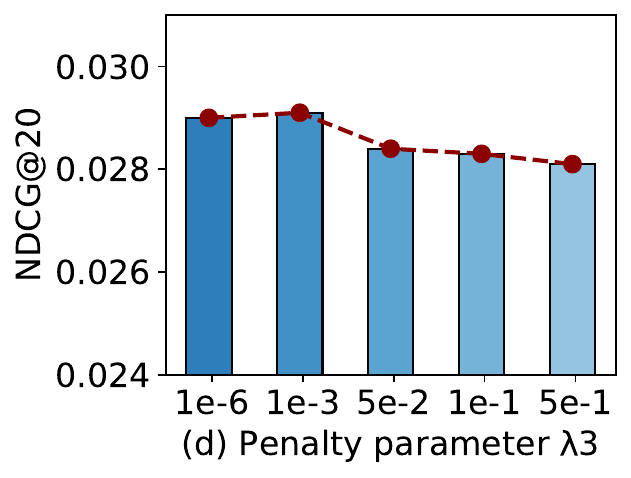}}
	\end{minipage}
    \begin{center}
     \end{center}
     \vspace{-0.4cm}
	\caption{Evaluating the impact of model hyperparameter sets on recommendation performance.}
	\label{fig: hyper_analysis}
\end{figure*}
\subsection{Overall Performance (RQ1)}
Based on the results shown in Table~\ref{tab: main-results}, we derive the following key observations:
\textbf{(1) CNSDiff significantly outperforms existing SOTA heuristic negative sampling methods under distribution shift scenarios.}  
Specifically, when compared to the top-performing heuristic method, MixGCF, CNSDiff demonstrates a notable performance improvement across multiple evaluation metrics on both the KuaiRec and Food datasets, and also demonstrates substantial gains on the Yelp2018 dataset. These results validate the effectiveness of CNSDiff in enhancing the OOD generalization ability of recommendation models.  
We mainly attribute this significant performance gain to two fundamental components of CNSDiff:
\begin{itemize}[leftmargin=0pt]
    \item \textbf{Diffusion-based negative sample generation}, which breaks away from the limitations of traditional heuristic sampling methods that rely on pre-defined candidate sets. Instead, it allows sampling of negative items with varying hardness levels from the diffusion process, thus reducing the likelihood of false negative samples.
    \item \textbf{Causal Regularization}: This component explicitly mitigates the influence of spurious correlations introduced by environment-related confounders during model training. By incorporating a causal regularization term into the objective function, it effectively blocks the erroneous causal pathways through which false negative samples could otherwise reinforce and amplify such misleading correlations. As a result, the framework not only enhances the stability of the learned representations but also promotes better generalization by enforcing invariance across different environmental conditions
\end{itemize}
Moreover, we observe that heuristic negative sampling methods generally fail to maintain strong OOD generalization under diverse types of distribution shifts. This further supports our motivation: \textit{heuristic strategies tend to introduce false hard negatives, which mislead the model into learning spurious patterns and consequently harm its generalization performance.}
\textbf{(2) CNSDiff also consistently outperforms existing SOTA OOD generalization methods and diffusion-based baselines.}  
Concretely, CNSDiff achieves an average performance gain of 13.96\% across multiple datasets. We attribute this improvement to its unique use of \textbf{causal regularization during the negative sampling phase}.  
Unlike prior methods, CNSDiff explicitly models both the latent environment confounders and the negative sample generation process, enabling it to (i) mitigate the misleading effects of unobserved environment variables and (ii) synthesize high-quality negative samples with varying levels of robustness. This mechanism compels the model to learn more accurate and robust decision boundaries, which in turn leads to stronger generalization performance under distribution shifts.

\begin{table}[t]
\renewcommand{\arraystretch}{1.2}
\setlength{\tabcolsep}{0.6mm}
\caption{Ablation Study on Three Datasets.}
\begin{tabular}{c|c|cccc}
\hline
Dataset & Method & R@10 & N@20 & R@20 & N@20 \\
\hline
\multirow{4}{*}{Food} 
& CNSDiff w/o Mixup & 0.0291 & 0.0218 & 0.0480 & 0.0285 \\
& CNSDiff w/o Align. & 0.0280 & 0.0217 & 0.0482 & 0.0284 \\
& CNSDiff w/o Causal & 0.0290 & 0.0212  &  0.0480 &  0.0286 \\
& CNSDiff w/o All &0.0275 & 0.0201 & 0.0463 & 0.0279\\
\rowcolor{gray!30}  & CNSDiff & \textbf{0.0300} & \textbf{0.0222} & \textbf{0.0490} & \textbf{0.0300} \\
\hline
\multirow{4}{*}{Yelp2018} 
& CNSDiff w/o Mixup & 0.0020 & 0.0012& 0.0033 & 0.0016\\
& CNSDiff w/o Align. & 0.0018 & 0.0011 & 0.0025 & 0.0014 \\
& CNSDiff w/o Causal & 0.0018 &  0.0011& 0.0028  &  0.0015 \\
& CNSDiff w/o All & 0.0015 & 0.0005& 0.0018& 0.0010\\
\rowcolor{gray!30} & CNSDiff  & \textbf{0.0101} & \textbf{0.0069} & \textbf{0.0163} & \textbf{0.0091} \\
\hline
\multirow{4}{*}{KuaiRec} 
& CNSDiff w/o Mixup & 0.0892 & 0.5460 & 0.1329 & 0.4697 \\
& CNSDiff w/o Align. & 0.0798  & 0.5299 & 0.1225 &  0.4569 \\
& CNSDiff w/o Causal & 0.0888 & 0.5518 & 0.1340 &  0.4681 \\
& CNSDiff w/o All & 0.0760 & 0.5078 & 0.1174  & 0.4300\\
\rowcolor{gray!30} &  CNSDiff & \textbf{0.0909} & \textbf{0.5572} & \textbf{0.1384} & \textbf{0.4781}\\
\hline
\end{tabular}
\label{tab: ab_test}
\end{table}

\subsection{Ablation Study}
This section presents ablation studies to validate the effectiveness of each module in CNSDiff. Specifically,  
\textbf{CNSDiff w/o Mixup} removes the interpolation between diffusion-generated and randomly sampled negatives.  
\textbf{CNSDiff w/o Align.} eliminates the embedding alignment between generated negatives and original items.  
\textbf{CNSDiff w/o Causal} disables the causal regularization designed to mitigate the influence of unobserved confounders.  
\textbf{CNSDiff w/o All} removes all three components simultaneously. The results are summarized in Table~\ref{tab: ab_test}.
We observe the following insights:
\textbf{(1)} Removing the Mixup module consistently degrades performance across datasets, confirming its role in enhancing training robustness by increasing sample diversity.
\textbf{(2)} Disabling embedding alignment leads to even greater performance drops in most cases, indicating that maintaining semantic consistency between generated and real items is vital for learning reliable decision boundaries.
\textbf{(3)} Omitting causal regularization also results in performance decline, validating its effectiveness in suppressing spurious correlations and improving generalization under distribution shifts.
\textbf{(4)} The removal of all three modules leads to the worst overall performance, demonstrating that Mixup, alignment, and causal regularization function complementarily and are jointly critical to CNSDiff's success.
These results collectively underscore the necessity of each component in achieving strong performance, particularly in OOD scenarios.

\begin{figure*}[t]
	\centering
	\begin{minipage}{0.22\linewidth}
		\centerline{\includegraphics[width=\textwidth]{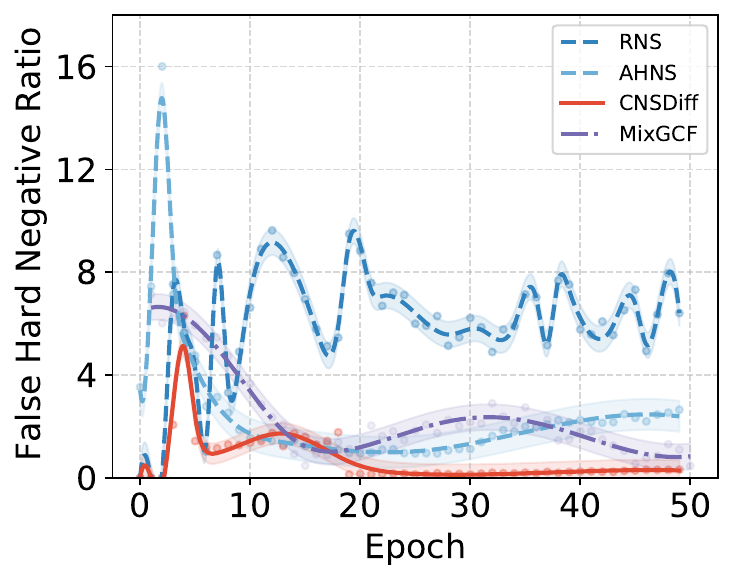}}
		\centerline{\footnotesize\;\;\;\;\;\;\;\; (a) Food Dataset: False Negative Rate}
	\end{minipage}
	\begin{minipage}{0.22\linewidth}
		\centerline{\includegraphics[width=\textwidth]{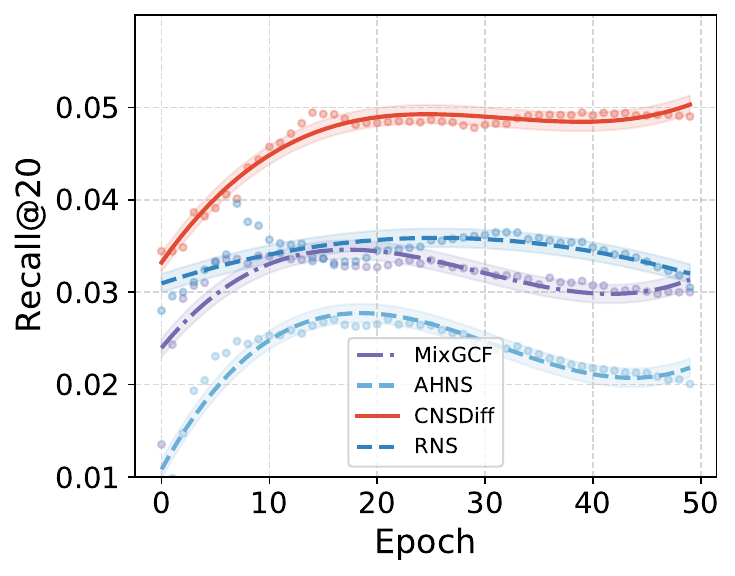}}
		\centerline{\footnotesize \;\;\;\;\;\;\;\;\;\;\;\;(b) Food Dataset: Recall@20}
	\end{minipage}
	\begin{minipage}{0.22\linewidth}
		\centerline{\includegraphics[width=\textwidth]{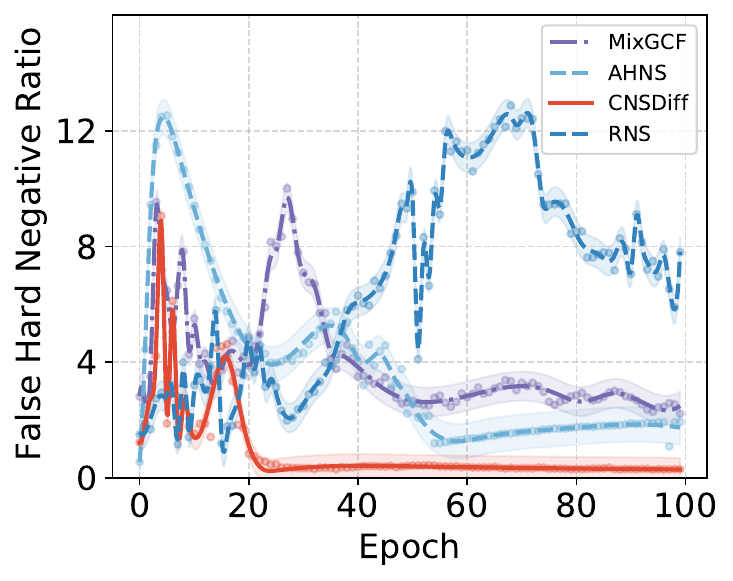}}
        \centerline{\footnotesize\;\;\;\;\;\;\; (c) Yelp Dataset: False Negative Rate}
	\end{minipage}
    \begin{minipage}{0.22\linewidth}
		\centerline{\includegraphics[width=\textwidth]{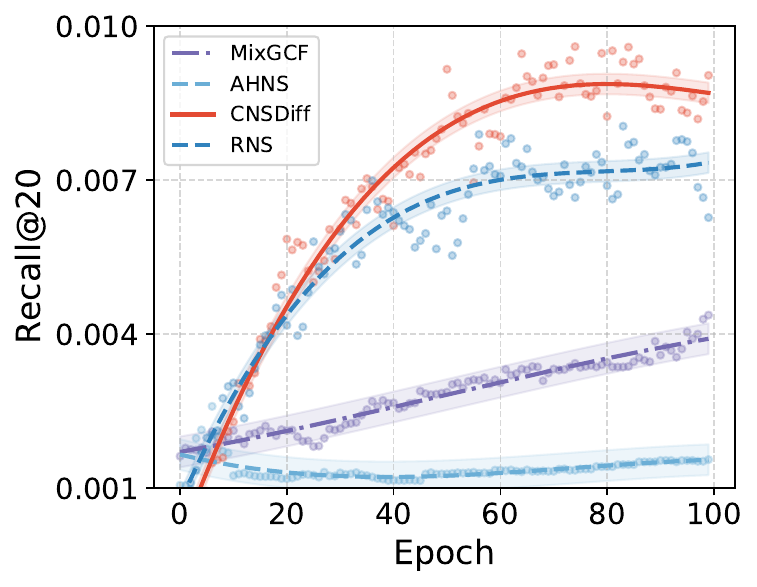}}
    \centerline{\footnotesize\;\;\;\;\;\;\;\;\;\;\;\; (d) Yelp Dataset: Recall@20}
	\end{minipage}
     \vspace{-0.3cm}
	\caption{False Negative Rate and Recall@20 Metric Over Epochs for Different-based Negative sampling.}
	\label{fig: false_neg}
    \vspace{-0.4cm}
\end{figure*}

\subsection{Hyperparameter Analysis (RQ3)}
To thoroughly evaluate the robustness and sensitivity of our proposed model, we conduct a comprehensive ablation study on several key hyperparameters, as illustrated in Figure~\ref{fig: hyper_analysis}. Our key analyses and conclusions are as follows:
\begin{itemize}[leftmargin=0pt]
    \item \textbf{Diffusion step $T$:} This parameter controls how far the sampling process explores in the latent space. We observe that the performance initially improves with larger $T$, peaking around $T=20$, and then gradually degrades. This suggests that a moderate diffusion depth enables the model to generate sufficiently diverse yet semantically meaningful negatives. However, excessive diffusion (e.g., 
    $T$>100) leads to noisy or overly distant samples, which impede learning and increase computational overhead.
    \item \textbf{Number of negative samples:} The model maintains stable performance across a range of negative sample counts, with optimal results achieved when using 5 to 10 negatives per positive instance. This indicates that incorporating multiple hard negative views benefits learning, though the marginal gain saturates as the number increases further. However, an excessive number of negative samples will also increase computational costs. In practice, a balance must be struck between computational overhead and performance.
    \item \textbf{ Regularization coefficient $\lambda_3$ (for diffusion sampling):} This penalty controls the semantic difficulty of sampled negatives. Small values (e.g., $1\mathrm{e}{-6}$ to $1\mathrm{e}{-3}$) yield better performance, while large penalties (e.g., $\geq 5\mathrm{e}{-2}$) significantly hurt NDCG. This highlights the importance of balancing diversity and semantic consistency.
    \item \textbf{Regularization coefficient $\lambda_3$}. We observe a similar trend: moderate regularization improves generalization, while overly strong constraints reduce embedding expressiveness. Mild semantic alignment helps enforce meaningful representations without overfitting.
\end{itemize}
Furthermore, in Table~\ref{tab: alpha_beta_ablation}, we further investigate the impact of different choices of alpha and beta on model performance.
Overall, Best performance is achieved with a diffusion step $T \approx 20$, using 5--10 negatives per positive sample, and applying mild regularization on both sampling and contrastive alignment losses.

\subsection{False Negative Analysis of CNSDiff (RQ4)}
To evaluate the effectiveness of CNSDiff in mitigating FHNS, we measure the proportion of false negatives among generated negative samples at each epoch. A sample is considered a false negative if its cosine similarity with any test item exceeds 0.99. Experiments are conducted on the Food and Yelp datasets, and the results are shown in Figure~\ref{fig: false_neg}. We summarize the key observations as follows:
\textbf{(1)} CNSDiff substantially reduces the false hard negative ratio compared to heuristic sampling methods. This advantage stems from its diffusion-based generation mechanism, which avoids relying on fixed candidate pools that often include near-duplicate positives. Notably, the false negative ratio is relatively high in the early training stages due to the predominance of random sampling before the diffusion-based negatives are fully integrated.
\textbf{(2)} CNSDiff achieves faster convergence and better performance, further validating the effectiveness of the causal regularization term. By suppressing spurious correlations introduced by FHNS, the model gains stronger generalization ability under distribution shifts.

\section{Related Work}

\subsection{Negative Sampling in Recommendation}
Negative sampling is a widely adopted technique for training recommendation models~\cite{ma2024negative,madushanka2024negative}. Over the past decade, various strategies have been proposed to improve training efficiency and performance, such as random negative sampling~\cite{rendle2012bpr,he2017neural,zhao2025symmetric,zhang2023asymmetrical}, popularity-based sampling~\cite{zhang2013optimizing,zhu2021popularity}, and predefined sampling methods~\cite{zhang2024practical,zheng2018drn}. Most of these fall under static strategies, where negative samples remain fixed during training, limiting the ability to mine informative hard negatives.
To address this, dynamic sampling approaches have emerged that select semantically similar negatives from candidate pools~\cite{yang2020understanding,ding2020simplify,chen2023revisiting,chen2023fairly,ma2023exploring}, interpolate between positives to generate harder negatives~\cite{huang2021mixgcf,lai2024adaptive}, or synthesize high-quality samples via generative models like GANs~\cite{guo2020ipgan}. With the rise of large language models (LLMs), recent work has explored LLM-generated semantic negatives to enhance generalization~\cite{zhao2025can}.
Distinct from these methods, our work revisits the negative sampling paradigm by highlighting the risk of false negatives introduced by heuristic sampling. Such samples can amplify spurious correlations unrelated to user preference, thereby degrading model generalization under distribution shift.

\subsection{Diffusion-based Recommendation}
Owing to their powerful generative and representation learning capabilities, diffusion models have achieved significant success across a range of generative tasks~\cite{croitoru2023diffusion,cao2024survey,xing2024survey}, spurring the development of various diffusion-based recommendation approaches~\cite{wei2025diffusion,lin2024survey}. Existing works can be broadly categorized into three directions:
\textbf{Data augmentation and robust representation learning}: Diffusion models are used to enhance robustness via noise-injection and denoising, benefiting tasks like sequential recommendation~\cite{liu2023diffusion,wu2023diff4rec,jiang2024diffkg,ma2024plug}, graph-based modeling~\cite{zhao2025distributionally,zhao2025graph}, and social network denoising~\cite{li2024recdiff,li2025dual,zhao2024denoising}. \textbf{Generative modeling core}: Some works adopt diffusion models as the primary mechanism for simulating user-item interactions, enabling direct modeling of preferences or rankings~\cite{wang2023diffusion,liu2024preference}. \textbf{Personalized content generation}: Diffusion models are also used to generate user-conditioned content, such as personalized images or product styles in creative recommendation settings~\cite{xu2024diffusion,lu2021personalized}.
Different from above, we explore a novel perspective that leverages diffusion models to synthesize negative samples of varying hardness. Our goal is to mitigate the impact of confounders and enhance both training effectiveness and out-of-distribution generalization for ID-based recommendation models. In Appendix~\ref{ap: distinction}, we further discuss the distinctions between our approach and existing diffusion-based recommendation methods.

\section{Conclusion} 
In this work, we propose CNSDiff, a causal negative sampling framework leveraging diffusion models to improve recommendation robustness under distribution shifts. Unlike traditional sampling methods prone to false negatives, CNSDiff synthesizes negatives in the latent space with causal regularization and hardness control, mitigating spurious correlations and enhancing out-of-distribution generalization.
Extensive experiments on three real-world datasets confirm CNSDiff’s consistent superiority over strong baselines.Ablation studies further validate the effectiveness of the proposed components. This work highlights the potential of integrating generative modeling and causal inference for robust recommendation. Future work will explore theoretical foundations and extensions to multi-modal and online recommendation.

\bibliographystyle{ACM-Reference-Format}
\bibliography{reference}

\appendix
\section{Proof and Derivation}
\subsection{The Proof of Theorem 3.1}
\label{pf: T_3.1}

We analyze the generalization gap between the training loss under corrupted labels and the test loss under true labels. Let the training and test objectives be defined as:
\begin{align}
\mathcal{L}_{\text{train}}(f_\theta) &= \mathbb{E}_{(X,\tilde{Y}) \sim P_{\text{train}}}[\ell(f_\theta(X), \tilde{Y})], \\
\mathcal{L}_{\text{test}}(f_\theta) &= \mathbb{E}_{(X,Y) \sim P_{\text{test}}}[\ell(f_\theta(X), Y)].
\end{align}
Assume the noisy label $\tilde{Y}$ is generated via environment-dependent false negative process:
\begin{equation}
P(\tilde{Y} = 1 \mid X) = \mathbb{E}_{E}\left[P(Y = 1 \mid X, E)(1 - \eta(E))\right],
\end{equation}
where $\eta(E) := P(\tilde{Y} = 0 \mid Y = 1, E)$ is the false negative rate conditioned on environment variable $E$.
Suppose the model takes the form $f_\theta(X) = g(X) + \lambda h(z_E)$, where $z_E$ is a latent representation influenced by $E$ (e.g., item popularity). Using a first-order Taylor expansion, we approximate:
\begin{equation}
\ell(f_\theta(X), \tilde{Y}) \approx \ell(f_\theta(X), Y) - \eta(E)P(Y = 1 \mid X, E) \cdot \frac{\partial \ell}{\partial f_\theta}.
\end{equation}
Taking expectation over $P_{\text{train}}$, we have:
\begin{align}
\mathcal{L}_{\text{train}}(f_\theta) &\approx \mathbb{E}_{(X,E) \sim P_{\text{train}}} \left[ \ell(f_\theta(X), Y) \right] \nonumber \\
&\quad - \mathbb{E}_{(X,E) \sim P_{\text{train}}} \left[ \eta(E)P(Y = 1 \mid X, E) \cdot \frac{\partial \ell}{\partial f_\theta} \right].
\end{align}
Hence, the generalization gap becomes:
\begin{align}
\mathcal{L}_{\text{test}}(f_\theta) - \mathcal{L}_{\text{train}}(f_\theta) &\approx \mathbb{E}_{(X,E) \sim P_{\text{test}}}[\ell(f_\theta(X), Y)] \nonumber \\
&\quad - \mathbb{E}_{(X,E) \sim P_{\text{train}}}[\ell(f_\theta(X), Y)] \nonumber \\
&\quad + \mathbb{E}_{(X,E) \sim P_{\text{train}}}\left[\eta(E)P(Y = 1 \mid X, E) \cdot \frac{\partial \ell}{\partial f_\theta}\right].
\end{align}
We focus on the environment-induced shift. Note that:
\begin{equation}
\mathbb{E}_{P_{\text{test}}}[f_\theta(X)] - \mathbb{E}_{P_{\text{train}}}[f_\theta(X)] = \lambda \cdot \left( \mathbb{E}_{P_{\text{test}}}[h(z_E)] - \mathbb{E}_{P_{\text{train}}}[h(z_E)] \right).
\end{equation}
Assuming $h$ is $L_h$-Lipschitz, the total variation inequality yields:
\begin{equation}
\left| \mathbb{E}_{P_{\text{test}}}[h(z_E)] - \mathbb{E}_{P_{\text{train}}}[h(z_E)] \right| \leq L_h \cdot \mathrm{TV}(P_{\text{train}}(E), P_{\text{test}}(E)).
\end{equation}
We define the environment sensitivity of $f_\theta$ as:
\begin{equation}
\gamma := \mathbb{E}_{X} \left[ \left\| \frac{\partial f_\theta(X)}{\partial z_E} \right\| \right] \cdot \mathrm{Var}_E(\eta(E)).
\end{equation}
Combining all terms, we obtain the following lower bound on the generalization gap:
\begin{equation}
\mathcal{L}_{\text{test}}(f_\theta) - \mathcal{L}_{\text{train}}(f_\theta) \geq \gamma \cdot \mathrm{TV}(P_{\text{train}}(E), P_{\text{test}}(E)) + \varepsilon_{\text{inv}},
\end{equation}
where $\varepsilon_{\text{inv}}$ denotes the residual irreducible error due to noise and approximation limits.

\subsection{Proof of Theorem 3.2: Variational Lower Bound for Deconfounded Learning Objective}
\label{proof：Variational}
We aim to derive a variational lower bound for the deconfounded learning objective $\log p_\theta(z_T | \text{do}(z_0), E)$, where $z_0$ is the initial embedding, $z_T$ is the negative sample generated via a diffusion model, and $E$ is the environment with discrete states $e$. The diffusion model is defined by the transition probabilities $p_\theta(z^{(l+1)} | z^{(l)}, e)$, and the goal is to learn causal relationships robust to distribution shifts.

\subsection*{Do-Calculus Rules}
We use two fundamental rules of do-calculus (Pearl et al., 2016) for a causal DAG with nodes $Z_0$, $Z_T$, and $E$:
\begin{enumerate}
    \item \textbf{Action/Observation Exchange}: 
    \[
    P(z_T | \text{do}(z_0), \text{do}(e)) = P(z_T | \text{do}(z_0), e), \quad \text{if } (Z_T \perp E | Z_0)_{A_{Z_0E}}
    \]
    \item \textbf{Insertion/Deletion of Actions}: 
    \[
    P(z_T | \text{do}(z_0), \text{do}(e)) = P(z_T | \text{do}(z_0)), \quad \text{if } (Z_T \perp E | Z_0)_{A_{Z_0E}}
    \]
\end{enumerate}

\subsection*{Derivation of the Deconfounded Objective}
The deconfounded objective is:
\[
p_\theta(z_T | \text{do}(z_0), E) = \sum_e p_\theta(z_T | \text{do}(z_0), e, E) p(e | \text{do}(z_0))
\]
Using the do-calculus rules:
\begin{itemize}
    \item By the action/observation exchange rule, assuming $(Z_T \perp E | Z_0, e)_{A_{Z_0E}}$:
    \[
    p_\theta(z_T | \text{do}(z_0), e, E) = p_\theta(z_T | z_0, e, E)
    \]
    \item By the insertion/deletion rule, assuming $(E \perp Z_0)_{A_{Z_0}}$:
    \[
    p(e | \text{do}(z_0)) = p_0(e)
    \]
\end{itemize}
Thus:
\[
p_\theta(z_T | \text{do}(z_0), E) = \sum_e p_\theta(z_T | z_0, e, E) p_0(e) = \mathbb{E}_{p_0(e)} \left[ p_\theta(z_T | z_0, e, E) \right]
\]
For the diffusion model, we have:
\[
p_\theta(z_T | z_0, e, E) = \prod_{l=0}^{T-1} p_\theta(z_{t+1} | z_{t}, e)
\]
The log-likelihood is:
\[
\log p_\theta(z_T | \text{do}(z_0), E) = \log \sum_{z_{1}, \dots, z_{T-1}} \sum_e \prod_{l=0}^{T-1} p_\theta(z_{t+1} | z_{t}, e) p_0(e)
\]

\subsection*{Variational Lower Bound}
To derive the lower bound, we introduce a variational distribution $q_\phi(e \mid z_t)$ to approximate the posterior of $e$. We split the derivation into intermediate steps ($t = 0, \dots, T{-}2$) and the final step ($t = T{-}1$).

\subsubsection*{Intermediate Steps ($t = 0, \dots, T{-}2$)}
For each term:
\[
\log \sum_{z_{t+1}} \sum_e p_\theta(z_{t+1} \mid z_t, e) \, p_0(e)
\]
Rewrite using $q_\phi(e \mid z_t)$:
\[
\log \sum_{z_{t+1}} \sum_e \left[ p_\theta(z_{t+1} \mid z_t, e) \, p_0(e) \, \frac{q_\phi(e \mid z_t)}{q_\phi(e \mid z_t)} \right]
\]
Applying Jensen's inequality:
\[
\geq \sum_e q_\phi(e \mid z_t) \cdot \log \left[ \sum_{z_{t+1}} \frac{p_\theta(z_{t+1} \mid z_t, e) \, p_0(e)}{q_\phi(e \mid z_t)} \right]
\]
\[
= \mathbb{E}_{q_\phi(e \mid z_t)} \left[ \log \frac{p_\theta(z_{t+1} \mid z_t, e) \, p_0(e)}{q_\phi(e \mid z_t)} \right]_{z_{t+1}}
\]
where we use $\sum_{z_{t+1}} p_\theta(z_{t+1} \mid z_t, e) = 1$.

\subsubsection*{Final Step ($t = T{-}1$)}
For the final step:
\[
\log \sum_e p_\theta(z_T \mid z_{T-1}, e) \, p_0(e)
\]
\[
= \log \sum_e \left[ p_\theta(z_T \mid z_{T-1}, e) \, p_0(e) \, \frac{q_\phi(e \mid z_{T-1})}{q_\phi(e \mid z_{T-1})} \right]
\]
\[
\geq \mathbb{E}_{q_\phi(e \mid z_{T-1})} \left[ \log \frac{p_\theta(z_T \mid z_{T-1}, e) \, p_0(e)}{q_\phi(e \mid z_{T-1})} \right]
\]

\subsubsection*{Combined Result}
Combining the results, we obtain:
\[
\log p_\theta(z_T \mid \text{do}(z_0), E) \geq \sum_{t=0}^{T-1} \mathbb{E}_{q_\phi(e \mid z_t)} \left[ \log \frac{p_\theta(z_{t+1} \mid z_t, e) \, p_0(e)}{q_\phi(e \mid z_t)} \right]
\]
The marginalization over $z_1, \dots, z_{T-1}$ is implicit through the delta predictive distribution $p_\theta(z_{t+1} \mid z_t, e)$.

\subsection*{Equality Condition}
Equality holds in Jensen's inequality if the function inside the expectation is constant with respect to $e$. For each step:
\[
p_\theta(z^{(l+1)} | z^{(l)}, e) p_0(e) = c \cdot q_\phi(e | z^{(l)})
\]
Summing over $e$:
\[
\sum_{z^{(l+1)}} \sum_e p_\theta(z^{(l+1)} | z^{(l)}, e) p_0(e) = c \cdot \sum_e q_\phi(e | z^{(l)}) = c
\]
Thus:
\[
q_\phi(e | z^{(l)}) = \frac{p_\theta(z^{(l+1)} | z^{(l)}, e) p_0(e)}{\sum_{z^{(l+1)}} \sum_{e'} p_\theta(z^{(l+1)} | z^{(l)}, e') p_0(e')}
\]
Since:
\[
p(e | z^{(l+1)}, z^{(l)}) = \frac{p_\theta(z^{(l+1)} | z^{(l)}, e) p(e | z^{(l)})}{\sum_{e'} p_\theta(z^{(l+1)} | z^{(l)}, e') p(e' | z^{(l)})}
\]
Assuming $p(e | z^{(l)}) = p_0(e)$, we have:
\[
q_\phi(e | z^{(l)}) = p(e | z^{(l+1)}, z^{(l)}) \cdot \frac{p_0(e)}{p(e | z^{(l)})}
\]

\subsection{Derivation of the Causal Regularization Term.}
\label{proof: Causal}
Eq.~(17) can be further reformulated into the following global variational form:
\[
\log p_\theta(z_T \mid \text{do}(z_0), E = e)
\]
 \[
 \geq \mathbb{E}_{q_\phi(e)} \left[ \sum_{t=0}^{T-1} \log p_\theta(z_{t+1} \mid z_t, e) + \log \frac{p_\theta(e \mid \text{do}(z_0))}{q_\phi(e)} \right].
 \]
The second term inside the expectation can be decomposed as:
\[
\mathbb{E}_{q_\phi(e)} \left[ \log \frac{p_\theta(e \mid \text{do}(z_0))}{q_\phi(e)} \right] 
= - D_{\mathrm{KL}} \left( q_\phi(e) \,\|\, p_\theta(e \mid \text{do}(z_0)) \right),
\]
which naturally introduces a KL divergence as a global regularization term enforcing alignment between the variational posterior and the interventional distribution over environment variables.
In practice, since the diffusion model follows a Markov process, we only require pairwise transitions $ (z_{t-1}, z_t) $ and the final state $ z_T $. Therefore, the variational approximation can be locally applied as:
\[
D_{\mathrm{KL}}\left(q_\phi(z_t \mid z_{t-1}, e) \,\|\, p_\theta(z_t \mid z_{t-1}, e)\right),
\]
which ensures consistency between the learned transition dynamics and the generative model at each step.
Furthermore, the term 
\[
\mathbb{E}_{q_\phi(e)} \left[ \log p_\theta(z_T, e \mid \text{do}(z_0)) \right]
\]
in the ELBO can be decomposed into:
\[
\sum_{t=0}^{T-1} \mathbb{E}_{q_\phi(e)} \left[ \log p_\theta(z_{t+1} \mid z_t, e) \right] 
+ \mathbb{E}_{q_\phi(e)} \left[ \log p_\theta(e \mid \text{do}(z_0)) \right],
\]
which enables a practical approximation of the interventional log-likelihood $ \log p_\theta(z_T \mid \text{do}(z_0), E) $, while simultaneously minimizing the global KL divergence:
\[
D_{\mathrm{KL}}\left(q_\phi(e) \,\|\, p_\theta(e \mid z_T, \text{do}(z_0))\right).
\]
This formulation unifies both the generative reconstruction loss and causal regularization into a coherent variational framework for robust, deconfounded learning.
\begin{figure}[t]
	\centering
	\begin{minipage}{0.48\linewidth}
		\centerline{\includegraphics[width=\textwidth]{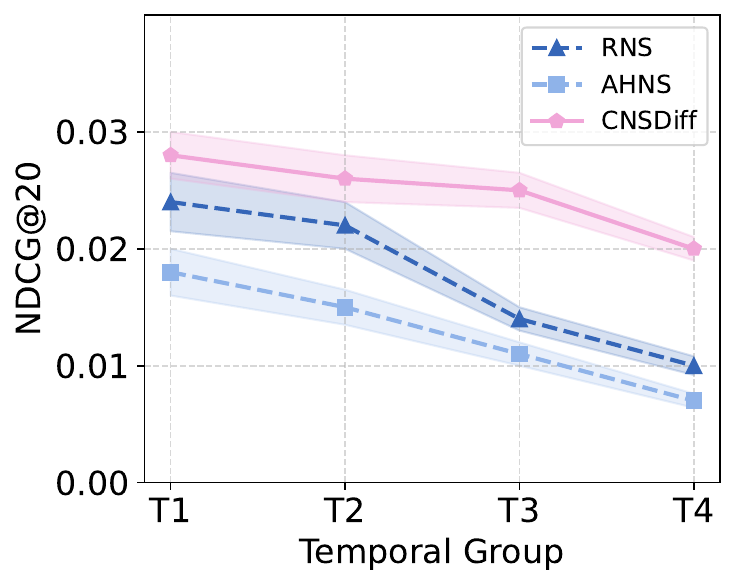}}
		\centerline{\footnotesize\;\;\;\;\;\;\;\; (a) }
	\end{minipage}
	\begin{minipage}{0.48\linewidth}
		\centerline{\includegraphics[width=\textwidth]{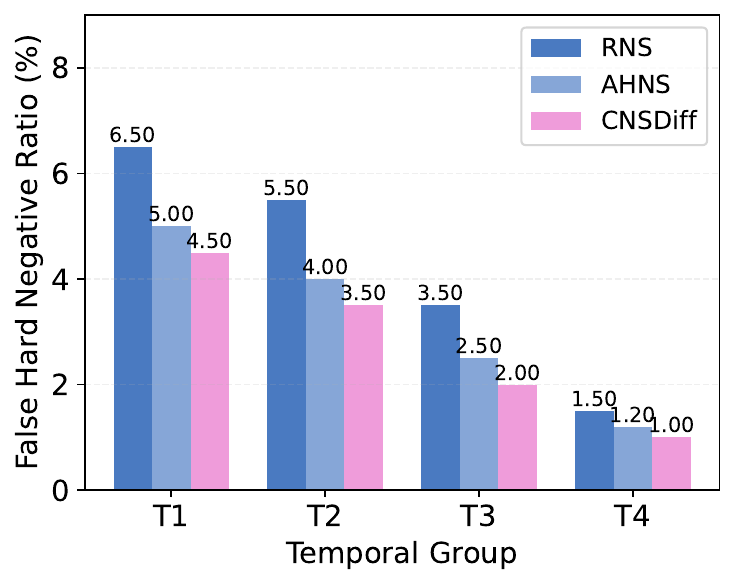}}
		\centerline{\footnotesize \;\;\;\;\;\;\;\;\;\;\;\;(b) }
	\end{minipage}
     \vspace{-0.3cm}
	\caption{Performance and false hard negative ratio Across different temporal groups.}
	\label{fig: empirical_study_temporal}
\end{figure}
\begin{algorithm}[t]
\caption{Training Procedure of CNSDiff}
\label{alg:CNSDiff}
\textbf{Input:} User-item interaction data $\{(u, v)\}$, model hyperparameters \\
\textbf{Output:} Trained model parameters $\theta$
\begin{algorithmic}[1]
\FOR{each training epoch}
    \STATE Compute user embeddings $u$ and item embeddings $z_0$ via GCN; treat $z_0$ as the initial state for diffusion
    \STATE Construct diffusion input set $\mathcal{D}_t$ by sampling negatives with varying hardness levels via Eq.~(14)
    \STATE Select hardest negative $e_h$ with maximum inner product with $u$ from $\mathcal{D}_{h_t}$ via Eq.~(20)
    \STATE Mix $e_h$ with a random negative $e_r$ to generate synthetic negative
    \STATE Align the synthetic negative to the item semantic space via Eq.~(22)
    \STATE Compute the total loss $\mathcal{L}_{\text{total}}$ via Eq.~(23)
    \STATE Update model parameters: $\theta \leftarrow \theta - \eta \cdot \nabla_\theta \mathcal{L}_{\text{total}}$
\ENDFOR
\end{algorithmic}
\end{algorithm}

\label{ap: dataset_detail}
\begin{table}[t]
\centering
 \caption{Detailed statistics for each dataset. For each negative sampling method, we conduct 50 epochs per dataset and measure the average runtime per epoch.}
 \renewcommand{\arraystretch}{1} 
    \begin{tabular}{c|ccccl}
     \hline
    \rowcolor{gray!30} Dataset & \#Users  & \#Items & \#Interactions & Density \\ 
    \hline
    Food & 7,809 &  6,309 & 216,407 & $4.4\times10^{-3}$\\
   KuaiRec & 7,175 & 10,611 & 1,153,797 & $1.5\times10^{-3}$  \\
    Yelp2018 & 8,090 & 13,878 & 398,216 & $3.5\times 10^{-3}$ \\
    \hline
    \end{tabular}
    \label{tab: dataset}
\end{table}

\begin{table}[t]
\centering
\caption{Running Time Comparison.}
\begin{tabular}{c|cccc}
\hline
\rowcolor{gray!30} Datasets & RNS & AHNS & MixGCF &CNSDiff \\
\hline
Food & 0.41 s & 4.28 s & 4.98 s & 2.74 s\\
Yelp2018 & 1.01 s & 8.72 s &  10.01 s & 7.04 s s \\
KuaiRec & 3.04 s & 36.98 s & 31.06 s & 19.58 s\\
\hline
\end{tabular}
\label{tab: time}
\end{table}

\begin{table*}[t]
\centering
\caption{Performance under different initial and final values of $\alpha$ and $\beta$}
\begin{tabular}{cccc|cccc}
\toprule
\textbf{Dataset} & \textbf{Initial [$\alpha$, $\beta$]} & \textbf{Final [$\alpha$, $\beta$]} 
& & \textbf{R@10} & \textbf{N@10} & \textbf{R@20} & \textbf{N@20} \\
\midrule
\multirow{3}{*}{\textbf{Yelp2018}} 
& [0.1, 9.9] & [1, 9]     & & 0.0105 & 0.0069 & 0.0163 & 0.0091 \\
& [1, 9]      & [5, 5]     & & 0.0099 & 0.0068 & 0.0163 & 0.0091 \\
& [2, 8]      & [9, 1]     & & 0.097 & 0.0065 & 0.0161 & 0.0090 \\
\midrule
\multirow{3}{*}{\textbf{KuaiRec}} 
& [0.1, 9.9]  & [9.9, 0.1] & & 0.0900 & 0.5516 & 0.1357 & 0.4673 \\
& [1, 9]      & [5, 5]     & & 0.0891 & 0.5530 & 0.1388 & 0.4688 \\
& [2, 8]      & [9, 1]     & & 0.0909 & 0.5572 & 0.1384 & 0.4678 \\
\midrule
\multirow{3}{*}{\textbf{Food}} 
& [0, 10]     & [10, 0]    & & 0.0304 & 0.0219 & 0.0497 & 0.0287 \\
& [1, 9]      & [5, 5]     & & 0.0301 & 0.0217 & 0.0493 & 0.0285 \\
& [2, 8]      & [9, 1]     & & 0.0304 & 0.0225 & 0.0496 & 0.0290 \\
\bottomrule
\end{tabular}
\label{tab: alpha_beta_ablation}
\end{table*}

\begin{table}[htbp]
\centering
\caption{Running Time Comparison.}
\begin{tabular}{c|cccc}
\hline
\rowcolor{gray!30} Datasets & RNS & AHNS & MixGCF &CNSDiff \\
\hline
Food & 0.41 s & 4.28 s & 4.98 s & 2.74 s\\
Yelp2018 & 1.01 s & 8.72 s &  10.01 s & 7.04 s s \\
KuaiRec & 3.04 s & 36.98 s & 31.06 s & 19.58 s\\
\hline
\end{tabular}
\label{tab: time}
\end{table}

\section{Dataset and Processing Details}
\label{ap: dataset}
Table~\ref{tab: dataset} summarizes the key statistics of each dataset, including the number of users, items, interactions, and the corresponding sparsity levels. Below, we provide a brief introduction to each dataset:
\begin{itemize}[left=0pt]
    \item \textbf{Food}: This dataset contains over 230,000 recipes along with millions of user interactions, such as reviews and ratings. It is well-suited for studying user preferences and developing food-related recommendation systems.
    \item \textbf{KuaiRec}: Collected from the Kuaishou platform, this fully-observed dataset includes dense user-item interaction logs with minimal missing data, making it ideal for analyzing the effects of exposure bias and data sparsity in recommendation tasks.
    \item \textbf{Yelp2018}: Comprising user reviews, ratings, and business metadata, this dataset is widely used for recommendation and user behavior modeling.
\end{itemize}

\textbf{Preprocessing Details}.  
To ensure data quality, we apply filtering criteria as follows:  
Users with fewer than 15 interactions are removed from the Food dataset, and users with fewer than 25 interactions are excluded from Yelp2018 . Items with fewer than 50 interactions are also filtered out from these datasets.  
Interactions rated 4 or higher are considered positive samples. For KuaiRec, we consider interactions with a watch ratio of 2 or greater as positive signals.
We construct three typical types of out-of-distribution (OOD) scenarios across these datasets:

\begin{itemize}[left=0pt]
    \item \textbf{Popularity Shift}: To simulate changes in item popularity, we randomly sample 20\% of interactions to form the OOD test set, ensuring a uniform popularity distribution. The remaining interactions are split into training, validation, and IID test sets in a 7:1:2 ratio. This setting is applied to Yelp2018.
    \item \textbf{Temporal Shift}: To capture temporal dynamics, we sort interactions chronologically for each user and select the most recent 20\% as the OOD test set. The remaining interactions are divided into training, validation, and IID test sets using a 7:1:2 split. This shift setting is applied to the Food dataset.
    \item \textbf{Exposure Shift}: For KuaiRec, the fully-exposed small matrix is used as the OOD test set, while the larger matrix collected from the online platform is split into training, validation, and IID test sets (7:1:2). This setting captures the distribution shift caused by differing exposure mechanisms.
\end{itemize}

\section{Hyperparameters Settings}
\label{ap: Hyper}
We implement our CNSDiff model using PyTorch, and all experiments are conducted on a single NVIDIA RTX 6000 GPU with 48 GB of memory.  
Following the default settings of the baselines, we set the embedding dimension to 64 and the number of convolution layers to 32 for all models to ensure a fair comparison.  
For MixGCF, DNS(M,N), and AHNS, the size of the candidate pool for negative sampling is set to 32. We tune the hyperparameters of CNSDiff as follows:
\begin{itemize}[left=0pt]
    \item The number of steps $T$: [10, 20,50,100,200]
    \item Regularization coefficient $\lambda_2$: [1e-6, 1e-3, 5e-2, 1e-1, 5e-1 ]
    \item Regularization coefficient $\lambda_3$: [1e-6, 1e-3, 5e-2, 1e-1, 5e-1 
\end{itemize}
We report the impact of different combinations of $\alpha$ and $\beta$ on model performance in Table~5. In general, the model achieves optimal performance when using the combinations of $[\alpha, \beta] = [2, 8]$ and $[9, 1]$.

\section{Baselines}
We compare CNSDiff with three types of baseline methods in terms of performance.

\noindent \textbf{Negative sampling methods}.
\begin{itemize}[left=0pt]
    \item \textbf{RNS} \cite{koren2009matrix} is a simple yet efficient baseline that randomly selects unobserved user-item pairs as negatives.
    \item \textbf{DNS (M, N)} \cite{shi2023theories} dynamically selects the hardest $M$ negatives from $N$ randomly sampled candidates to better approximate top-$K$ ranking objectives and enhance training effectiveness.
    \item \textbf{MixGCF} \cite{huang2021mixgcf} enhances collaborative filtering by mixing multiple graph convolutional paths with different propagation ranges, enabling more expressive user and item representations and improving recommendation performance.
    \item \textbf{AHNS}~\cite{lai2024adaptive} adaptively selects negative samples with appropriate hardness levels during training, effectively mitigating false positive and false negative issues caused by fixed-hardness sampling strategies and improving recommendation performance.
    \item \textbf{DMNS} ~\cite{nguyen2024diffusion}introduces a conditional diffusion-based multi-level negative sampling strategy that generates negative nodes with controllable hardness levels from the latent space, enabling more effective contrastive learning for graph link prediction.
\end{itemize}

\noindent \textbf{ Distribution shift-aware methods}. 
\begin{itemize}[left=0pt]
\item \textbf{CDR} ~\cite{wang2023causal}: Employs a temporal variational autoencoder to capture preference shifts and learns sparse influences from various environments.
\item \textbf{InvCF}~\cite{zhang2023invariant}: Targets the popularity bias by learning disentangled representations of user preferences and item popularity, without relying on predefined popularity distributions.

\item \textbf{AdvInfoNCE}~\cite{zhang2024empowering}: An enhanced contrastive learning method that incorporates hardness-aware ranking criteria to improve the model's generalization in recommendation.

\item \textbf{DR-GNN}~\cite{wang2024distributionally}: A graph neural network-based approach for out-of-distribution recommendation, leveraging distributionally robust optimization to handle data distribution shifts.

\end{itemize}

\noindent \textbf{ Diffusion-based methods}.
\begin{itemize}[left=0pt]
    \item \textbf{DiffRec}~\cite{wang2023diffusion}: A diffusion-based recommendation framework that learns robust user and item representations by simulating preference propagation through a stochastic diffusion process on the user-item graph.
    \item \textbf{HDRM}~\cite{yuan2025hyperbolic}: introduces a hyperbolic diffusion framework for recommendation, leveraging the geometric properties of hyperbolic space to preserve anisotropic and directional structures, enhancing the expressiveness of latent representations.
    \item \textbf{CausalDiff}~\cite{zhao2025graph} mitigates the impact of environmental confounders by integrating causal inference with diffusion-based representation learning, enabling the model to learn environment-invariant user-item representations for OOD recommendation.
\end{itemize}

\section{Distinction from Existing Diffusion-based Recommendation Methods}
\label{ap: distinction}
In this section, we further discuss the connections and distinctions between \textbf{CNSDiff} and existing \textit{diffusion-based recommendation methods}, highlighting the unique advantages of our proposed framework.
\subsection{Connections to Diffusion-based Methods}
Diffusion models have recently been explored in recommendation systems for robust representation learning~\cite{zhao2025distributionally,zhao2025graph}, graph denoising~\cite{zhao2024denoising}, and sequential modeling~\cite{li2023diffurec, ma2024plug}. Similar to these works, \textbf{CNSDiff} employs a conditional diffusion process to synthesize representations in the latent space. However, unlike models such as DiffRec~\cite{wu2023diff4rec} or HDRM~\cite{nguyen2024diffusion}, which focus on improving user/item embeddings or interaction modeling, CNSDiff is the first to leverage diffusion \textit{explicitly for negative sample generation}, with controllable hardness via diffusion steps.
\subsection{Key Differences and Innovations}
Compared to prior diffusion-based methods, \textbf{CNSDiff} introduces the following key innovations:
\begin{itemize}[left=0pt]
    \item \textbf{Targeted Purpose:} Existing methods typically apply diffusion for embedding enhancement or data augmentation. In contrast, CNSDiff utilizes diffusion as a generative mechanism for constructing \textit{informative and hardness-controllable} negative samples, aligning more directly with the training objective.
    \item \textbf{Causal Regularization:} CNSDiff uniquely incorporates a causal regularization term based on the backdoor criterion, which explicitly mitigates spurious correlations induced by latent environmental confounders—a factor ignored by conventional diffusion-based models.
    \item \textbf{FHNS Mitigation:} Unlike prior methods that may still suffer from false hard negatives (FHNS) under distribution shifts, CNSDiff not only generates negatives from a principled latent distribution but also enforces \textit{environment-invariant generation}, enhancing robustness and generalization.
\end{itemize}
\subsection{Empirical Advantages}
As shown in Table~\ref{tab: main-results}, CNSDiff consistently outperforms other diffusion-based baselines, including DiffRec, HDRM, and CausalDiff, across three real-world datasets. Specifically:
\begin{itemize}[left=0pt]
    \item CNSDiff achieves an \textbf{average performance gain of 13.96\%} in Recall and NDCG metrics over state-of-the-art diffusion-based methods.
    \item Figure~\ref{fig: false_neg} shows that CNSDiff maintains a \textbf{lower false hard negative ratio} throughout training, leading to faster convergence and more robust generalization.
\end{itemize}
These results demonstrate that simply applying diffusion modeling is insufficient under distribution shift. Instead, \textbf{explicitly modeling causal structures and mitigating confounding effects is essential} for robust recommendation in the wild.

\end{document}